\documentclass[letterpaper, 10 pt, journal, twoside]{IEEEtran}
\usepackage{times}
\usepackage{epsfig}
\usepackage{graphicx}
\usepackage{amsmath}
\usepackage{amssymb}
\usepackage{algpseudocode}
\usepackage{algorithm}
\usepackage{algorithmicx}
\usepackage{array}
\usepackage{booktabs}
\usepackage{cite}
\usepackage{multirow}
\usepackage{soul}
\usepackage{subcaption}
\usepackage{widetable}
\usepackage{xcolor}
\usepackage[normalem]{ulem}
\usepackage[pagebackref=true,breaklinks=true,letterpaper=true,colorlinks,bookmarks=false]{hyperref}
\newcommand{\etal}{\textit{et al}.}
\newcommand{\ie}{\textit{i}.\textit{e}.}
\newcommand{\eg}{\textit{e}.\textit{g}.}
\newcommand{\xhdr}[1]{\vspace{4pt} \noindent {\textbf{#1}}}

\IEEEoverridecommandlockouts                              

\begin{document}

\title{
Stepwise Goal-Driven Networks for Trajectory Prediction
}

\author{
    Chuhua Wang$^{1}$, Yuchen Wang$^{1}$, Mingze Xu$^{2}$, David J. Crandall$^{1}$
    \thanks{Manuscript received September 9, 2021; revised November 23, 2021; accepted December 30, 2021. This paper was recommended for publication by Editor Cesar Cadena Lerma upon evaluation of the Associate Editor and Reviewers’ comments. This work was supported in part by a grant from the U.S. Navy (N00164-21-1-1002) and the Indiana University Office of the Vice Provost for Research through the Emerging Areas of Research project ``Learning: Brains, Machines, and Children.'' \textit{(Chuhua Wang and
    Yuchen Wang contributed equally to this work.)}  \textit{(Corresponding author: Chuhua Wang.)} }
    \thanks{$^{1}$The authors are with the Luddy School of Informatics, Computing, and Engineering, Indiana University, Bloomington, IN 47408, USA (e-mail: cw234@indiana.edu; wang617@indiana.edu; djcran@indiana.edu).}
    \thanks{$^{2}$The author was with the Luddy School of Informatics, Computing, and Engineering, Indiana University, Bloomington, IN 47408, USA. He is now with Amazon/AWS AI, Seattle, WA, USA (e-mail: mx6@indiana.edu).}
    \thanks{Digital Object Identifier (DOI): see top of this page.
}
}
\markboth{IEEE ROBOTICS AND AUTOMATION LETTERS. PREPRINT VERSION. ACCEPTED January, 2022}
{WANG \MakeLowercase{\textit{et al.}}: STEPWISE GOAL-DRIVEN NETWORKS FOR TRAJECTORY PREDICTION}

\maketitle

\begin{abstract}
We propose to predict the future trajectories of observed agents (\eg, pedestrians or vehicles) by estimating and using their goals at multiple time scales. We argue that the goal of a moving agent may change over time, and modeling goals continuously provides more accurate and detailed information for future trajectory estimation. To this end, we present a recurrent network for trajectory prediction, called Stepwise Goal-Driven Network (SGNet). Unlike prior work that models only a single, long-term goal, SGNet estimates and uses goals at multiple temporal scales. In particular, it incorporates an encoder that captures historical information, a stepwise goal estimator that predicts successive goals into the future, and a decoder that predicts future trajectory. We evaluate our model on three first-person traffic datasets (HEV-I, JAAD, and PIE) as well as on three bird's eye view datasets (NuScenes, ETH, and UCY), and show that our model achieves state-of-the-art results on all datasets.
Code has been made available at: \textit{\url{https://github.com/ChuhuaW/SGNet.pytorch}}.
\end{abstract}


\vspace{-5pt}
\section{Introduction}

\IEEEPARstart{P}{redicting} the future behavior of other agents
is crucial in the real world~\cite{rudenko2020human}:
safe driving requires predicting future movements
of other cars and pedestrians, for example, while effective social interactions require
anticipating the actions of social partners.
However, predicting another agent's future actions is challenging because such actions depend on numerous factors including the environment and the other agent's internal state
(\eg, its intentions and goals).
Recent work~\cite{rhinehart2019precog,mangalam2020not,zhao2020tnt,yao2020bitrap}
has explored goal-driven methods for trajectory prediction,
which explicitly try to estimate the other agent's long-term goal to 
help predict its future behavior.
While these models make a significant step forward, they 
adopt the simplistic assumption
that an agent's intentions are exclusively represented by a single long-term goal.

However, work in psychology and cognitive science
suggests that people base their actions not on a single long-term goal,
but instead on a series of goals at different time scales.
Some literature~\cite{locke1990theory,10.2307/258611} uses the term \textit{intention} to refer to a representation of planned actions
and the term \textit{goal} to stress the end result of an action. This suggests that
a series of goals may better represent an intention than using
a single, long-term goal.
Besides, people's
intentions build up progressively, and each decision made in the past may
have an impact on how the future is determined. 
For example,
a pedestrian plans a trajectory before
crossing the street; when he or she takes a step forward, the prior
planned trajectory and the actual situation are both taken into
account to update a new set of stepwise goals.
We show that estimating stepwise goals at the initial time step and carrying them over to subsequent time steps results in a more precise model of intention and better guidance for future trajectory.
\begin{figure}[t]
    \centering
    \includegraphics[trim={0cm 13cm 0 10cm},clip,width=0.97\linewidth]{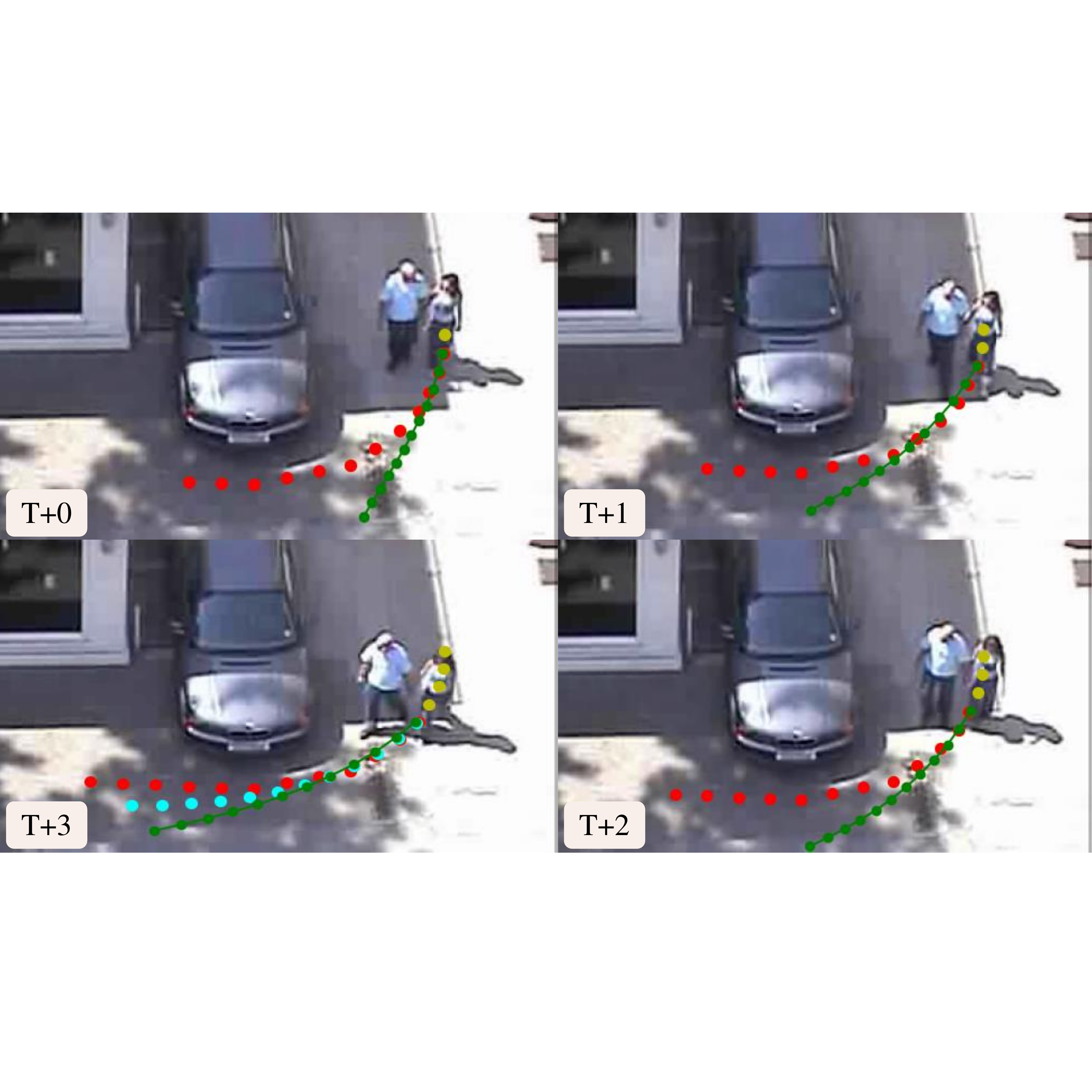}
    \vspace{-3pt}
    \caption{Illustration of stepwise goals (clockwise from top left). Panes (T+0),
    (T+1), and (T+2) show the initial, second, and third time step,
    respectively, where yellow indicates observed trajectory, red
    indicates ground truth future trajectory, green indicates stepwise
    goals, and cyan indicates prediction. As the model accumulates
    historical data, goals build up progressively and become more
    accurate, creating better representations of intention. The goals help
    predict future trajectory in Pane (T+3). Goals are represented as
    feature vectors, but we map them to locations for visualization
    purposes.
}
\vspace{-10pt}
\label{fig:teaser}
\vspace{-9pt}
\end{figure}

We present Stepwise Goal-Driven Network
(SGNet) to address the trajectory prediction problem.
SGNet consists of three main components: (1) A stepwise goal estimator (SGE)
that predicts coarse future goals at multiple temporal scales to encode a comprehensive
representation of
the intention. To determine the significance of each stepwise goal, a
lightweight module with an attention mechanism is used.
(2) An
encoder that records past data in conjunction with predicted
stepwise goals, to incorporate a richer hidden representation that aids in predicting the future and in creating new stepwise goals for the next time step.
(3) A decoder that takes
advantage of stepwise goals to predict future trajectories. 
We show how stepwise goals evolve and aid in predicting future trajectory in Fig.~\ref{fig:teaser}.

We evaluate our model on multiple first- and third-person
datasets, including both vehicles and pedestrians, and compare to an
extensive range of existing work ranging from deterministic to
stochastic approaches. We surpass or match the state-of-the-art
performance on multiple benchmarks,
including different viewpoints (\ie, first- and third-person) and agents (\ie, cars and pedestrians).

The contributions of this paper are three-fold.
First, our work highlights a new direction 
for goal-driven trajectory prediction by modeling goals at multiple time scales. SGE is a versatile module that may be applied to a wide range of architectures. 
Second, we show how
to effectively incorporate each series of stepwise goals into an encoder and decoder.
By integrating stepwise objectives into the decoder, we can direct the trajectory prediction in the current step, and by embedding stepwise goals into encoder, we can generate more accurate future goals for the next time step.
Finally, our goal aggregator employs an attention mechanism to adaptively learn the relative relevance of each stepwise goal, thereby improving performance even further.

\vspace{-3pt}
\section{Related Work}

\vspace{-5pt}
\xhdr{Trajectory prediction from first-person views}
jointly models the motion of observed objects and the ego-camera.
Bhattacharyya~\etal~\cite{bhattacharyya2018long} propose the Bayesian LSTMs
to model observation uncertainty
and predict the distribution of future locations.
Yagi~\etal~\cite{yagi2018future} use multi-modal data,
such as human pose, scale, and ego-motion, as cues in a convolution-deconvolution (Conv1D)
framework to predict future pedestrian locations.
Yao~\etal~\cite{yao2018egocentric_icra}
introduce a multi-stream encoder-decoder 
that separately captures both object location and appearance.
Makansi~\etal~\cite{makansi2020multimodal}
estimate a reachability prior for 
objects from the semantic map and propagate them into the future.

\begin{figure*}[t]
\vspace{5pt}
\centering
\includegraphics[width=0.97\linewidth]{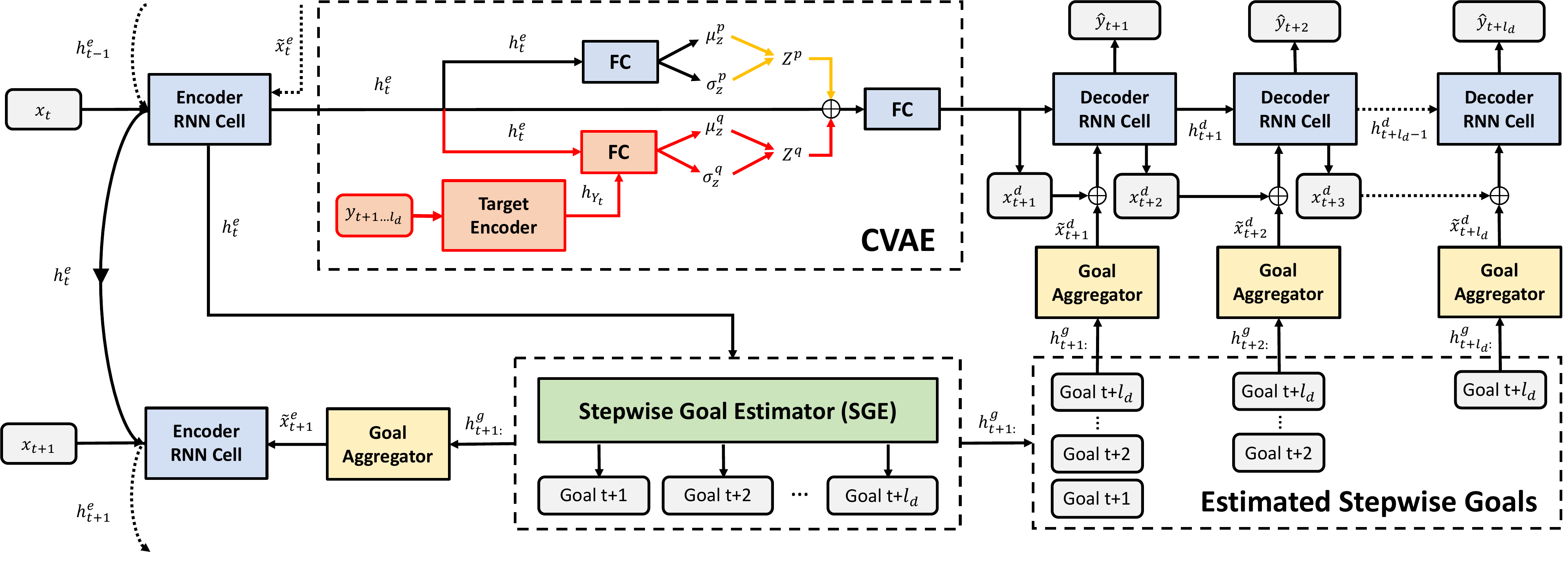}
\vspace{-5pt}
\caption{Visualization of SGNet. Arrows in red,
  yellow, and black indicate connections
  during training, inference, and both training and inference,
  respectively. Encoder time evolves vertically, from time  $t$ to $t + 1$, while decoder time flows horizontally, predicting the trajectory at $t+1$ to $t+l_d$. For deterministic results, we replace CVAE with a non-linear embedding.}
\vspace{-10pt}
\label{fig:network}
\end{figure*}

\xhdr{Trajectory prediction from a bird's eye view} simplifies the
problem by removing the ego-motion.
Alahi~\etal~\cite{alahi2016social} propose Social-LSTM to model
pedestrians' trajectories and interactions.  Their social
pooling module was improved by~\cite{gupta2018social} to capture
global context.  SoPhie~\cite{sadeghian2019sophie} applies generative
models to model the uncertainty
of future paths.  Lee~\etal~\cite{lee2017desire} use RNNs
with conditional variational autoencoders
(CVAEs) to generate multi-modal predictions.
Recent work~\cite{ivanovic2019trajectron,tang2019multiple,NEURIPS2019_d09bf415}
also proposes graph-based recurrent models, simultaneously predicting
potential trajectories of multiple objects,
while~\cite{salzmann2020trajectron++} exploits more dynamic and
heterogeneous inputs.  PLOP~\cite{buhet2020plop} and
Argoverse~\cite{chang2019argoverse} use the ego trajectory in
a bird's eye view map.
Simaug and SMARTS~\cite{liang2020simaug,zhou2020smarts} take advantage of
simulation data to train the prediction model.
Others~\cite{zhang2019sr,deo2018multi,brito2020social,xiao2020diff,liang2019peeking,choi2020shared} have explored multimodal
inputs, such as
Lidar~\cite{rhinehart2018r2p2,choi2019drogon,shah2020liranet,zeng2019end},
to aid in trajectory
prediction.

\xhdr{Goal-driven trajectory prediction} incorporates
estimated future goals.
Rhinehart~\etal~\cite{rhinehart2019precog}
anticipate multi-modal semantic actions as the goal
and conduct conditional forecasting using imitative models.
Deo~\etal~\cite{deo2020trajectory} estimate goal states
and fuse the results with past trajectories 
using maximum entropy inverse reinforcement learning.
PECNet~\cite{mangalam2020not} infers distant destinations to
improve long-range trajectory prediction.
TNT~\cite{zhao2020tnt} decomposes the prediction task into three stages:
predicting potential target states, generating trajectory state sequences,
and estimating trajectory likelihoods.
BiTraP~\cite{yao2020bitrap} uses a bi-directional decoder
on the predicted goal to improve long-term trajectory prediction.

In contrast to the above methods that only estimate and use the
final goal (\ie, the destination), our model predicts goals
at multiple temporal scales and incorporates them into our encoder-decoder
framework using attentive mechanisms.
\vspace{-15pt}
\section{Stepwise Goal-Driven Network (SGNet)}
\vspace{-2pt}

At time step $t$, given an object's observed trajectory in the last $\ell_e$ steps,
$\mathbf{X}_t = \{ \mathbf{x}_{t-\ell_e+1}, \mathbf{x}_{t-\ell_e+2}, \dots, \mathbf{x}_{t} \}$,
where $\mathbf{x}_t$ includes its bounding box
(\ie,  centroid position and width and height in pixels)
and motion (\eg, optical flow, velocity, and acceleration), 
our goal is to predict its future positions
$\mathbf{Y}_t = \{ \mathbf{y}_{t+1}, \mathbf{y}_{t+2}, \dots, \mathbf{y}_{t+\ell_d} \}$
in the next $\ell_d$ frames.

\vspace{-2pt}
\subsection{Overview}
\vspace{-1pt}
Because a person's intentions are a representation of
planned actions, employing a single long-term goal to portray it is
quite limited. To give a more comprehensive intention representation and
improve the quality of trajectory prediction, we propose estimating numerous smaller goals
along the way and explicitly
including them at each decoder time step.
Furthermore, people regularly adjust and optimize their intentions, and previous intentions
can impact how they perceive the present, as well as aid in the development of new future plans. Thus historical goals can be treated as additional information in the encoder for embedding the present representation and to forecast future stepwise goals.
One simple way to merge the stepwise goals is through average pooling as in~\cite{onlineaction2019iccv}. However, we believe that at each time step, each individual goal has a different impact on the prediction.
Average pooling smooths out the goal features and hence the important goal features may not be identified. 
Therefore, we use an aggregator that
adaptively learns the importance of each stepwise goal with an attention mechanism. 
We propose a new
recurrent encoder-decoder architecture, Stepwise Goal-Driven
Network (SGNet), which predicts goals step-by-step to provide guidance
during trajectory prediction as well as supplementary features to
help predict new goals at the next time step.

Fig.~\ref{fig:network} presents an overview of SGNet. In particular, the stepwise goal estimator
(SGE) predicts an object's future locations from $t+1$ to $t+\ell_d$
as stepwise goals, and embeds them as input to the decoder
in an incremental manner to ensure trajectory prediction is guided
without receiving any redundant information.
SGE also fuses and feeds all predicted goals
into the encoder for the next time step, which, as our experiments will show,
helps encode a
better representation of the current
information, and help predicting new stepwise goals for the next time step.

\vspace{-2pt}
\subsection{Encoder} \label{Encoder}
\vspace{-1pt}
The encoder captures an agent's movement behavior as a latent
vector by embedding its
historical trajectory $\mathbf{X}_t$ using a single fully-connected
layer. If additional motion features (\eg, optical flow)
are available, they are also embedded
using a separate fully-connected layer and concatenated with the trajectory
representations.
The input feature $\mathbf{x}_t^e$ is then concatenated with
the aggregated goal information $\widetilde{\mathbf{x}}_t^e$ from the previous time step $t-1$,
and then the new hidden state $h_{t}^e$ is updated through a recurrent cell.
Hidden state $h_{t+1}^{e}$ and goals $\widetilde{\mathbf{x}}_{t+1}^e$ are both set to zero for the first time step. 
We next discuss how to obtain the aggregated goal input. 

\vspace{-2pt}
\subsection{Stepwise Goal Estimator (SGE)} \label{Goal}
\vspace{-1pt}
The main idea of SGE is to generate coarse
stepwise goals to assist trajectory prediction in a
coarse-to-fine manner. These goals also help the network to create a new stepwise goal for the next step.
Consequently, we design SGE to
predict and convey the predicted coarse stepwise goals to both encoder
and decoder.
For encoder, at each time step $t$, a
set of stepwise goals from $t-1$ is concatenated with the input to serve
as supplementary features, helping the encoder learn a more
discriminative representation of the sequence. Since inaccurate future
goals may mislead the prediction, we use a goal aggregator that
adaptively learns the importance of each stepwise goal by using
an attention mechanism. Meanwhile, for decoder, at each time step $t+i \ (i \in [1, \ell_d]) $, a subset of stepwise goals
$h_{t+i:}$ serves as coarse guidance to help trajectory
prediction, and a goal aggregator again gathers the selected goals.

We define a generic SGE module $f_{SGE}$ as,
\begin{align}
    h_{t+1:}^{g} &= f_{SGE}(ReLU(\mathbf{W}_{\gamma}^T h^e_t + \mathbf{b}_{\gamma})) \label{SGE}
\end{align}
where $h_{t+1:}^g$ is a sequence of stepwise goals, and $h_{t+1:}^g = \{h_{t+1}^g,h_{t+2}^g, \dots, h_{t+\ell_d}^g\}$. $h_{t}^e$ is the encoder hidden state at time $t$.
We found that SGE is open to different implementations
including recurrent, convolution, and fully-connected layers,
as will be described in Sec.~\ref{instant}.
To regularize SGE to generate goals that
contain precise future information, we regress the goal position
$\widehat{\mathbf{Y}}_t^g$ using the same regressor defined in
Sec.~\ref{Decoder} to minimize the distance between goal position and the
ground truth.

\begin{figure}[t]
\centering
\includegraphics[width=0.82\linewidth]{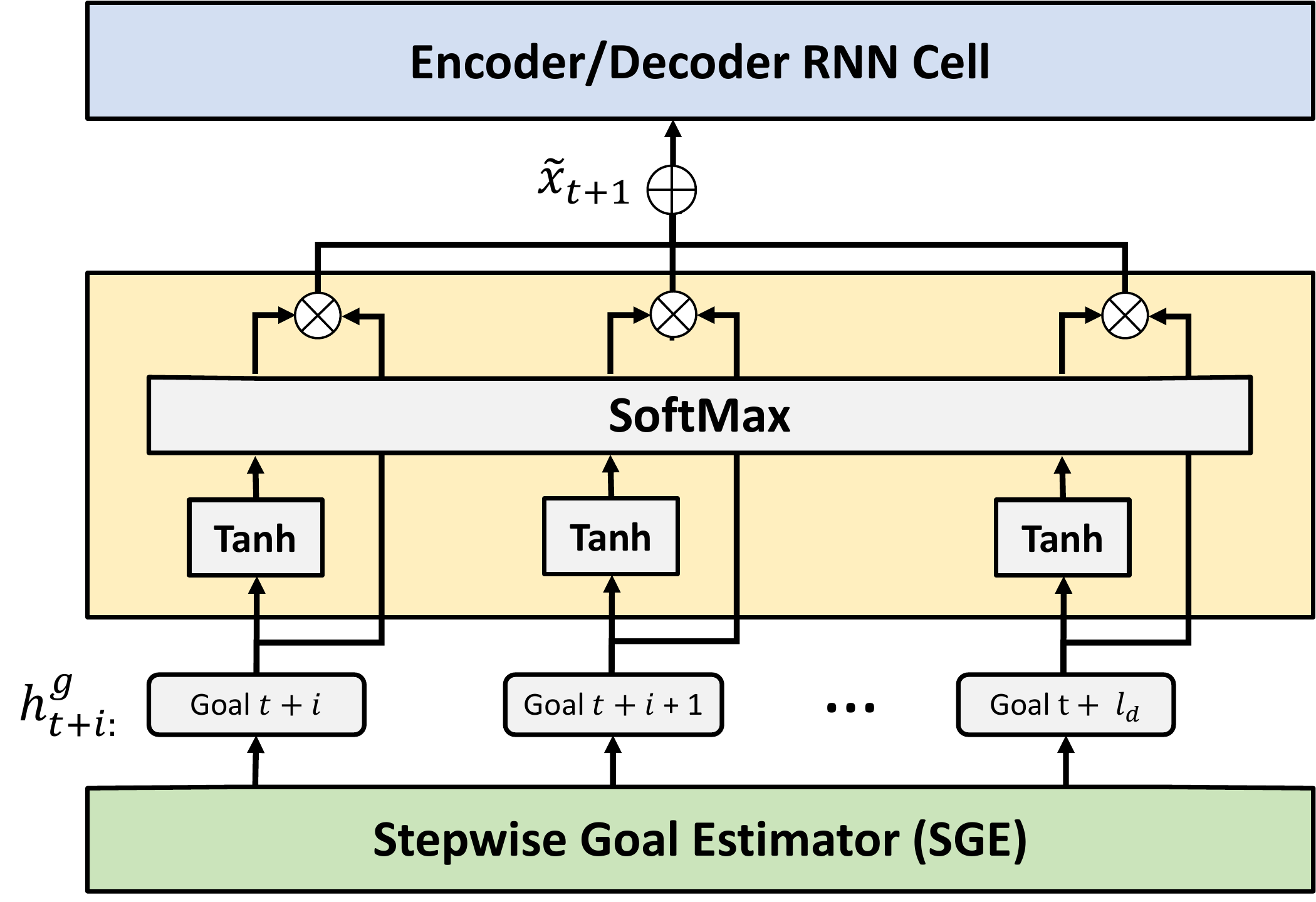}
\vspace{-2pt}
\caption{Detailed structure of Goal Aggregator. It receives  a set of stepwise goals as inputs and calculates the attention weights adaptively. The re-weighted goal features are fed into encoder or decoder.}
\vspace{-10pt}
\label{fig:attention}
\end{figure}

\xhdr{Goal Aggregator for Encoder and Decoder.}
We design goal aggregators for both the encoder and decoder, in order to combine and compress multiple goals into a single representation with attention. We define the goal aggregator,
\begin{align} \label{enc_goal_input}
    w &= \text{Softmax}(\mathbf{W}^T \text{Tanh}(h_{t+i:}^g) +\mathbf{b})\\
    \widetilde{\mathbf{x}}_{t+i} &= f_{attn}(h_{t+i:}^g) = \sum_{s=t+i}^{\ell_d} w_s h_s^g,
\end{align}
where $w$ is an attention vector corresponding to the probability distribution over a subset of
estimated goals, and $w_s$ is the weight for each individual goal $h_s^g$.
The overall structure is shown in Fig.~\ref{fig:attention}.

\textbf{For the encoder}, we hypothesize that the anticipation differs at each time step, and people's anticipation in the past may influence their perception of the present and future. As a result, at time $t$, the encoder receives stepwise goals $h_{t+1:}^g =
\{h_{t+1}^g, h_{t+2}^g, ..., h_{t+\ell_d}^g\}$. \textbf{For the decoder}, at
time step $t+i$, the decoder receives goals $h_{t+i:}^g =
\{h_{t+i}^g, h_{t+i+1}^g, ..., h_{t+\ell_d}^g\}$, and the goals before $t+i$ are ignored, 
because these are redundant and have been encoded in the historical information.

\vspace{-3pt}
\subsection{Conditional Variational Autoencoder (CVAE)}
\label{CVAE}
\vspace{-3pt}
A 
Conditional Variational Autoencoder (CVAE)
framework is applied to learn the distribution of
future trajectory $Y_t$ conditioned on the observed trajectory $X_t$ by introducing a latent variable $z$.  Following prior work~\cite{yao2020bitrap, mangalam2020not,bhattacharyya2018long}, our  CVAE consists of three components: recognition network $Q_\phi(z|X_t, Y_t)$, prior network
$P_\nu(z|X_t)$, and generation network $P_\theta(h_t^{d}|X_t, z)$,
where $\phi,\nu,\theta$ denote the parameters of these three networks,
and $h_t^{d}$ is the trajectory encoded by the generation
network.
Recognition, prior, and generation networks are implemented
with fully-connected layers.

\textbf{During training}, \label{CVAE_train} the ground truth future
trajectory $Y_t$ is fed into the target encoder to output the
hidden state $h_{Y_t}$. To capture the dependencies between observed
and ground truth trajectories, the recognition network takes
hidden states $h^e_t$ and $h_{Y_t}$ and predicts the distribution mean
$\mu_{z}^q$ and standard deviation $\sigma_{z}^q$. The prior network takes
$h^e_t$ only and predicts $\mu_{z}^p$ and
$\sigma_{z}^p$. 
We sample $z$ from $N(\mu_{z}^q,
\sigma_{z}^q)$ and concatenate it with $h^e_t$ to generate $h_t^{d}$
with the generation network. \textbf{During testing}, the ground truth
future trajectory is not available. The generation network concatenates $z$, which is sampled from $N(\mu_{z}^p,\sigma_{z}^p)$ with $h^e_t$ and produces $h_t^{d}$.

\vspace{-3pt}
\subsection{Decoder} \label{Decoder}
\vspace{-3pt}
Our recurrent decoder outputs the final trajectory $\mathbf{Y}_t = \{ \mathbf{y}_{t+1},\mathbf{y}_{t
+2},
\dots,  \mathbf{y}_{t+\ell_d}\}$ with a trajectory regressor. Given
$h_{t}^d$ and estimated goal input $\widetilde{\mathbf{x}}_{t+1}^d$, it produces a new hidden
state for the next time step through a recurrent cell.
Our trajectory regressor is a single fully-connected layer that takes hidden states $h_{t+i}^{d}$ and computes a trajectory
$\widehat{\mathbf{y}}_{t+i}$ at each time step,

\begin{table*}
\vspace{5pt}
\setlength\tabcolsep{4.9pt}
\small
\begin{center}
\begin{tabular}{ccccccccccc}
\toprule
\multicolumn{1}{l}{\multirow{4}{*}{SGE}} & \multicolumn{1}{l}{\multirow{4}{*}{Encoder}} & \multicolumn{1}{l}{\multirow{4}{*}{Decoder}} &\multicolumn{1}{l}{\multirow{4}{*}{\#Goals}} &\multicolumn{1}{l}{\multirow{4}{*}{Attention}} &\multicolumn{3}{c}{JAAD} & \multicolumn{3}{c}{PIE}\\
\cmidrule(lr){6-8} \cmidrule(lr){9-11}
\multicolumn{1}{c}{}& \multicolumn{1}{c}{}&\multicolumn{1}{c}{}&\multicolumn{1}{c}{}&\multicolumn{1}{c}{}& \multicolumn{1}{c}{\begin{tabular}[c]{@{}c@{}}MSE $\downarrow$\\
(0.5s / 1.0s / 1.5s)\end{tabular}} & \begin{tabular}[c]{@{}c@{}}C$_{MSE}$ $\downarrow$ \\
(1.5s)\end{tabular} & \begin{tabular}[c]{@{}c@{}}CF$_{MSE}$ $\downarrow$ \\(1.5s)\end{tabular} &
\multicolumn{1}{c}{\begin{tabular}[c]{@{}c@{}}MSE $\downarrow$ \\ (0.5s / 1.0s / 1.5s)\end{tabular}} &
\begin{tabular}[c]{@{}c@{}}C$_{MSE}$ $\downarrow$ \\ (1.5s)\end{tabular} &
\begin{tabular}[c]{@{}c@{}}CF$_{MSE}$ $\downarrow$ \\ (1.5s)\end{tabular} \\

\midrule

GRU & \checkmark &  & 45 & \checkmark &   90 / 372 / 1176 & 1117 & 4497 & 40 / 154 /  496  & 464 & 1939 \\
GRU &  & \checkmark & 45 & \checkmark &   87 / 350 / 1121 & 1065 & 4355 & 37 / 148 /  478  & 450 & 1891 \\
GRU & \checkmark & \checkmark & 45 &  & 85 / 341 / 1102 & 1044 & 4269 & 37 / 143 /  476  & 445 & 1905 \\
GRU & \checkmark & \checkmark & 1 & \checkmark & 94 / 372 / 1162 & 1077 & 4369 & 41 / 150 /  499  & 452 & 1919 \\
GRU & \checkmark & \checkmark & 2 & \checkmark & 87	/ 355 / 1140 & 1074 & 4394 & 40 / 148 /  482  & 449 & 1934 \\
GRU & \checkmark & \checkmark & 4 & \checkmark & 87 / 345 / 1100 & 1038 & 4248 & 38 / 144 /  479  & 448 & 1925 \\
GRU & \checkmark & \checkmark & 8 & \checkmark & 86 / 345 / 1088 & 1028 & 4179 & 38 / 145 /  476  & 446 & 1890 \\
GRU & \checkmark & \checkmark & 16 & \checkmark & 85 / 339 / 1082 & 1015 & 4173 & 38 / 145 /  473  & 443 & 1866 \\
GRU & \checkmark & \checkmark & 32 & \checkmark & 85 / 333 / 1064 & 1011 & 4146 & 37 / 142 /  467  & 438 & 1868 \\
MLP & \checkmark & \checkmark & 45 & \checkmark & 84 / 338 / 1098 & 1032 & 4376 & 39 / 153 / 496 & 464 & 1967\\
CNN & \checkmark & \checkmark & 45 & \checkmark & 88 / 344 / 1079 & 1016 & 4170& 37 / 145 / 470 & 441 & 1859\\
GRU & \checkmark & \checkmark & 45 & \checkmark & \textbf{82 / 328 / 1049} & \textbf{996} & \textbf{4076} & \textbf{34 / 133 / 442} & \textbf{413} & \textbf{1761}\\
\bottomrule
\end{tabular}
\end{center}
\vspace{-10pt}
\caption{Exploration study of our model on JAAD and PIE. $\downarrow$ denotes lower is better. The last row is our best model.}
\vspace{-5pt}
\label{tab:ablation_sge}
\end{table*}

\vspace{-3pt}
\subsection{Loss Functions}
\vspace{-2pt}
We use the Root Mean Square Error (RMSE) as loss function
to supervise trajectory prediction from our decoder.
For our stochastic model using CAVE, we follow~\cite{yao2020bitrap,bhattacharyya2018long} and adapt best-of-many (BoM) approach to minimize the distance
between our best prediction $\mathbf{\widehat{Y}}_{t}$ and target
$\mathbf{Y}_{t}$. This approach leads to more accurate and diverse
predictions and encourages the model to capture the true variation in data.
To ensure SGE to predict accurate goal states, we also optimize the prediction from SGE using RMSE between goal prediction
$\mathbf{\widehat{Y}}_{t}^g$ and ground truth $\mathbf{Y}_{t}$. Finally, we add KL-divergence loss (KLD) to optimize the prior
network in the CVAE. Thus, for each training sample, our final loss is summarized as follows,
\begin{align*}
    \mathcal{L}_{total} =  
    \min_{\forall k \in K} \mbox{RMSE}(\mathbf{\widehat{Y}}_{t}^k,\mathbf{Y}_{t}) 
    + \mbox{RMSE}(\mathbf{\widehat{Y}}_{t}^g,\mathbf{Y}_{t})\\
    + \; \mbox{KLD($Q_\phi(z|X_t, Y_t)$, $P_\nu(z|X_t)$)},
\end{align*}
where $\mathbf{\widetilde{Y}}_{t}^k$ is the $k$-th trajectory hypothesis from CVAE, $\mathbf{\widetilde{Y}}_{t}^g$ is the predicted stepwise goal location, and $\mathbf{Y}_{t}$ is the object's ground truth location at time $t$.

\vspace{-2pt}
\section{Experiments}
\vspace{-4pt}
\begin{table*}[t]
\small
\begin{center}
\begin{tabular}{lcccccc}
\toprule
\multicolumn{1}{l}{\multirow{3}{*}{Method}} & \multicolumn{3}{c}{JAAD} & \multicolumn{3}{c}{PIE}\\ \cmidrule(lr){2-4} \cmidrule(lr){5-7}
\multicolumn{1}{c}{}& \multicolumn{1}{c}{\begin{tabular}[c]{@{}c@{}}MSE $\downarrow$\\
(0.5s / 1.0s / 1.5s)\end{tabular}} & \begin{tabular}[c]{@{}c@{}}C$_{MSE}$ $\downarrow$ \\
(1.5s)\end{tabular} & \begin{tabular}[c]{@{}c@{}}CF$_{MSE}$ $\downarrow$ \\(1.5s)\end{tabular} &
\multicolumn{1}{c}{\begin{tabular}[c]{@{}c@{}}MSE $\downarrow$ \\ (0.5s / 1.0s / 1.5s)\end{tabular}} &
\begin{tabular}[c]{@{}c@{}}C$_{MSE}$ $\downarrow$ \\ (1.5s)\end{tabular} &
\begin{tabular}[c]{@{}c@{}}CF$_{MSE}$ $\downarrow$ \\ (1.5s)\end{tabular} \\

\midrule
Bayesian-LSTM~\cite{bhattacharyya2018long}     & 159 / 539 / 1535 & 1447 & 5615 &  101 / 296 /  855  & 811 & 3259 \\
FOL-X~\cite{yao2018egocentric}      & 147 / 484 / 1374 & 1290 & 4924 &  47 / 183 /  584  & 546 & 2303 \\
PIE$_{traj}$~\cite{Rasouli2019PIE}  & 110 / 399 / 1248 & 1183 & 4780 &  58 / 200 /  636  & 596 & 2477\\
BiTraP-D~\cite{yao2020bitrap}   &   93 / 378 / 1206 & 1105 & 4565 &  41 / 161 /  511  & 481 & 1949\\
\midrule 
SGNet-ED     &   \textbf{82} / \textbf{328} / \textbf{1049}  & \textbf{996} & \textbf{4076} &   \textbf{34} / \textbf{133} / \textbf{442}  & \textbf{413} & \textbf{1761} \\
\bottomrule
\end{tabular}
\end{center}
\vspace{-10pt}
\caption{Deterministic results on JAAD and PIE in terms of MSE/C$_{MSE}$/CF$_{MSE}$. $\downarrow$ denotes lower is better.}
\vspace{-10pt}
\label{tab:jaadpie_results}
\end{table*}

\begin{table*}[t]
\vspace{5pt}
\small
\begin{center}
\begin{tabular}{lcccccc}
\toprule
\multicolumn{1}{l}{\multirow{2}{*}{Method}} & \multicolumn{6}{c}{\begin{tabular}[c]{@{}c@{}}ADE (4.8s)~$\downarrow$ / FDE (4.8s)~$\downarrow$ \end{tabular}} \\
\cmidrule{2-7}    
\multicolumn{1}{c}{} & ETH   & HOTEL   & UNIV   & ZARA1   & ZARA2   & Avg \\
\midrule
Social-LSTM~\cite{alahi2016social} & 1.09 / 2.35 & 0.79 / 1.76 & 0.67 / 1.40 & 0.47 / 1.00 & 0.56 / 1.17 & 0.72 / 1.54 \\
Social-GAN~\cite{gupta2018social}  & 1.13 / 2.21 & 1.01 / 2.18 & 0.60 / 1.28 & 0.42 / 0.91 & 0.52 / 1.11 & 0.74 / 1.54 \\
MATF ~\cite{zhao2019multi}& 1.33 / 2.49 & 0.51 / 0.95 & 0.56 / 1.19 & 0.44 / 0.93 & 0.34 / 0.73 & 0.64 / 1.26 \\
FvTraj~\cite{bi2020fvtraj}   & 0.62 / 1.23 & 0.53 / 1.10 & 0.57 / 1.19 & 0.42 / 0.89 & 0.38 / 0.79 & 0.50 / 1.04 \\
STAR-D~\cite{YuMa2020Spatio}   & \textbf{0.56} / \textbf{1.11} & 0.26 / 0.50 & 0.52 / 1.15 & 0.41 / 0.90 & 0.31 / 0.71 & 0.41 / 0.87 \\
Trajectron++~\cite{salzmann2020trajectron++} & 0.71 / 1.68 & \textbf{0.22} / \textbf{0.46} & 0.41 / 1.07 & 0.30 / 0.77 & 0.23 / 0.59 & 0.37 / 0.91 \\
\midrule
SGNet-ED   & 0.63 / 1.38 & 0.27 / 0.63 & \textbf{0.40} / \textbf{0.96} & \textbf{0.26} / \textbf{0.64}  & \textbf{0.21} / \textbf{0.53} & \textbf{0.35} / \textbf{0.83} \\
\bottomrule
\end{tabular}
\end{center}
\vspace{-10pt}
\caption{Deterministic results on ETH and UCY in terms of ADE/FDE. $\downarrow$ denotes the lower the better.}
\vspace{-15pt}
\label{tab:eth_results}
\end{table*}

\subsection{Datasets}

\vspace{-5pt} \xhdr{First-person datasets}.
JAAD~\cite{kotseruba2016joint} and PIE~\cite{Rasouli2019PIE} have egocentric videos
recorded at 30 frames per second (fps), and consist of 2,800 and 1,835 pedestrian trajectories, respectively.
Following~\cite{Rasouli2019PIE}, we divided the datasets
into train (50\%), validation (10\%), and test (40\%) sets.
HEV-I~\cite{yao2019egocentric} includes 230 videos that are splitted into
40,000 train and 17,000 test samples.
Following~\cite{yao2018egocentric_icra}, the annotations are 
generated by using Mask-RCNN and Sort~\cite{bewley2016simple} with a Kalman filter.
We use 1.6 seconds of observations to predict future trajectories of length 0.5, 1.0, and 1.5 seconds.

\xhdr{Third-person dataset}.
ETH~\cite{pellegrini2009you} and UCY~\cite{lerner2007crowds} include 1,536 pedestrians in 5
sets of data with 4 unique scenes. Following prior
work~\cite{salzmann2020trajectron++}, a leave-one-out strategy is used
to split the train and test sets. We use 3.2 seconds
of observations to predict 4.8 seconds future trajectories.
NuScenes~\cite{nuscenes2019} is a dataset for autonomous driving, and we follow the their prediction challenge splits and settings. We use 2 seconds of observations to predict 6 seconds future trajectories.

\vspace{-3pt}
\subsection{Implementation Details}
\vspace{-1pt}
We use Gated Recurrent Units (GRUs) as
the backbone for both encoder and decoder with $512$ hidden size.
The length of the observation $\ell_e$ is
determined by the default setting of each benchmark: $\ell_e$ is $16$
on HEV-I, $15$ on JAAD, $15$ on PIE, and $8$ on ETH-UCY.  Object
bounding boxes are taken as inputs for JAAD and PIE, and
following~\cite{yao2018egocentric_icra}, optical flow is also included
on HEV-I.  We follow~\cite{salzmann2020trajectron++} to use object
centroids, velocities, and accelerations as inputs for ETH and
UCY.  We use the Adam~\cite{kingma2014adam} optimizer with
initial learning rate $5 \times 10^{-4}$,
which is dynamically reduced based on the validation loss.
Our models are optimized end-to-end with batch size $128$
and the training is terminated after $50$ epochs.

\vspace{-3pt}
\subsection{Evaluation Protocols}
\vspace{-1pt}
Our main evaluation metrics are \textbf{average displacement error
  (ADE)}, which measures accuracy along the whole trajectory, and
\textbf{final displacement error (FDE)}, which measures accuracy only
at the trajectory end point.  For first-person datasets, we use
upper-left and lower-right coordinates of bounding boxes to calculate
ADE and FDE, except for HEV-I where we only use the upper-left
coordinate to be consistent with prior work. In the bird's eye view
ETH and UCY datasets, we follow prior work to use the coordinates of
points.
To compare to the 
state-of-the-art~\cite{yao2018egocentric} in HEV-I,
we also use \textbf{final intersection over
  union (FIOU)}, 
the overlap between the predicted bounding box and ground truth at the
final step, which measures the model's ability to predict both
the scale and location of bounding boxes in the long-term.  We use \textbf{mean
  squared error (MSE)} to evaluate our performance on JAAD and
PIE, calculated according to the upper-left and lower-right
coordinates of the bounding box. 
\textbf{Center mean squared error (C$_{MSE}$)} and \textbf{center
  final mean squared error (CF$_{MSE}$)}
are similar to ADE and FED except 
they are computed based on the bounding box centroids.  All
results metrics used for HEV-I, JAAD, and PIE dataset are in pixels,
while for ETH and UCY 
we compute the ADE and FDE in Euclidean space.

\vspace{-3pt}
\subsection{Exploration Study}\label{exploration}
\vspace{-3pt}

We begin with experiments for design trade-offs and training
strategies of our architecture with JAAD and PIE.

\xhdr{How to implement SGE?} \label{instant}
We considered three instantiations of $f_{attn}$ in the SGE module. 
\textbf{First,} we implement SGE with GRUs (Table~\ref{tab:ablation_sge}, row 9).
The hidden size is set to 128, since we only need a lightweight module to
produce a coarse goal prediction. $h_{t+i}^g$ is defined to
be the  hidden state at time $t+i$, which is
initialized by using the encoder hidden state $h_t^e$ after a linear transformation.
The GRU input $\mathbf{x}_{t+1}^g$ is 
initialized with a zero vector
and updated using the auto-regressive strategy,
and the sequence of the output hidden state at each time step is used
as the stepwise goals.  \textbf{Second,} we try SGE with a multilayer
perceptron (MLP) (Table~\ref{tab:ablation_sge}, row 10) that takes the encoder hidden states $h_{t}^e$ as
input, and directly outputs $\ell_d$ goals of size  128. We call this SGNet-ED-MLP.  \textbf{Third,} we
follow~\cite{yagi2018future} and implement a convolution-deconvolution
framework for SGE to predict the stepwise goals (Table~\ref{tab:ablation_sge}, row 11).

As shown in
Table~\ref{tab:ablation_sge}, all of the above variants achieve the
state-of-the-art results, indicating that the SGE module is not sensitive to
the choice of $f_{attn}$.  The
results also suggest that using temporal models, such as GRU, as SGE is more effective and robust
in future trajectory prediction. Thus, we use this version in the remaining experiments (SGNet-ED).

\begin{table*}[t]
\vspace{5pt}
\small
\begin{center}
\begin{tabular}{lcccccc}
\toprule
\multicolumn{1}{l}{\multirow{3}{*}{Method (Best of 20)}} & \multicolumn{3}{c}{JAAD} & \multicolumn{3}{c}{PIE}\\ \cmidrule(lr){2-4} \cmidrule(lr){5-7}
\multicolumn{1}{c}{}& \multicolumn{1}{c}{\begin{tabular}[c]{@{}c@{}}MSE $\downarrow$\\
(0.5s / 1.0s / 1.5s)\end{tabular}} & \begin{tabular}[c]{@{}c@{}}C$_{MSE}$ $\downarrow$ \\
(1.5s)\end{tabular} & \begin{tabular}[c]{@{}c@{}}CF$_{MSE}$ $\downarrow$ \\(1.5s)\end{tabular} &
\multicolumn{1}{c}{\begin{tabular}[c]{@{}c@{}}MSE $\downarrow$ \\ (0.5s / 1.0s / 1.5s)\end{tabular}} &
\begin{tabular}[c]{@{}c@{}}C$_{MSE}$ $\downarrow$ \\ (1.5s)\end{tabular} &
\begin{tabular}[c]{@{}c@{}}CF$_{MSE}$ $\downarrow$ \\ (1.5s)\end{tabular} \\

\midrule
BiTrap-GMM~\cite{yao2020bitrap}& 153 / 250 / 585 & 501 & 998     & 38 / 90 / 209 & 171    & 368     \\
BiTrap-NP~\cite{yao2020bitrap}& 38 / 94 / 222   & 177    & 565     & 23 / 48 / 102 & 81     & 261     \\
SGNet-ED   & \textbf{37} / \textbf{86} / \textbf{197}   & \textbf{146}  & \textbf{443}     & \textbf{16} / \textbf{39} / \textbf{88}  & \textbf{66}     & \textbf{206}     \\
\bottomrule
\end{tabular}
\end{center}
\vspace{-10pt}
\caption{Stochastic results on JAAD and PIE in terms of MSE/C$_{MSE}$/CF$_{MSE}$. $\downarrow$ denotes lower is better.}
\vspace{-7pt}
\label{tab:jaadpie_cvae_results}
\end{table*}

\begin{table*}[t]
\small
\begin{center}
\begin{tabular}{lcccccc}
\toprule
\multicolumn{1}{l}{\multirow{2}{*}{Method (Best of 20)}} & \multicolumn{6}{c}{\begin{tabular}[c]{@{}c@{}}ADE (4.8s)~$\downarrow$ / FDE (4.8s)~$\downarrow$ \end{tabular}} \\
\cmidrule{2-7}    
\multicolumn{1}{c}{} & ETH   & HOTEL   & UNIV   & ZARA1   & ZARA2   & Avg \\
\midrule
Social-GAN~\cite{gupta2018social}  & 0.81 / 1.52 & 0.72 / 1.61 & 0.60 / 1.26 & 0.34 / 0.69 & 0.42 / 0.84 & 0.58 / 1.18 \\
Sophie~\cite{sadeghian2019sophie} & 0.70 / 1.43 & 0.76 / 1.67 & 0.54 / 1.24 & 0.30 / 0.63 & 0.38 / 0.78 & 0.54 / 1.15 \\
CGNS~\cite{li2019conditional}   & 0.62 / 1.40 & 0.70 / 0.93 & 0.48 / 1.22 & 0.32 / 0.59 & 0.35 / 0.71 & 0.49 / 0.97 \\
MATF GAN~\cite{zhao2019multi} & 1.01 / 1.75 & 0.43 / 0.80 & 0.44 / 0.91 & 0.26 / 0.45 & 0.26 / 0.57 & 0.48 / 0.90 \\
FvTraj~\cite{bi2020fvtraj}   & 0.56 / 1.14 & 0.28 / 0.55 & 0.52 / 1.12 & 0.37 / 0.78 & 0.32 / 0.68 & 0.41 / 0.85 \\
DSCMP~\cite{Tao_2020}   & 0.66 / 1.21 & 0.27 / 0.46 & 0.50 / 1.07 & 0.33 / 0.68 & 0.28 / 0.60 & 0.41 / 0.80 \\
PECNet~\cite{mangalam2020not}   & 0.54 / 0.87 & 0.18 / 0.24 & 0.35 / 0.60 & 0.22 / 0.39  & 0.17 / 0.30 & 0.29 / 0.48 \\
STAR~\cite{YuMa2020Spatio}   & 0.36 / \textbf{0.65} & 0.17 / 0.36 & 0.31 / 0.62 & 0.26 / 0.55 & 0.22 / 0.46 & 0.26 / 0.53 \\
Trajectron++~\cite{salzmann2020trajectron++} & 0.43 / 0.86 & \textbf{0.12} / \textbf{0.19} & 0.22 / 0.43 & 0.17 / 0.32 & 0.12 / 0.25 & 0.21 / 0.41 \\
BiTrap-GMM~\cite{yao2020bitrap} & 0.40 / 0.74 & 0.13 / 0.22 & 0.19 / 0.40 & 0.14 / 0.28 & 0.11 / 0.22 & 0.19 / 0.37 \\
BiTrap-NP~\cite{yao2020bitrap} & 0.37 / 0.69 & \textbf{0.12} / 0.21 & \textbf{0.17} / \textbf{0.37} & 0.13 / 0.29 & \textbf{0.10} / \textbf{0.21} & \textbf{0.18} / \textbf{0.35} \\
\midrule
SGNet-ED   & \textbf{0.35} / \textbf{0.65} & \textbf{0.12} / 0.24 & 0.20 / 0.42 & \textbf{0.12} / \textbf{0.24}  & \textbf{0.10} / \textbf{0.21} & \textbf{0.18} / \textbf{0.35} \\
\bottomrule
\end{tabular}
\end{center}
\vspace{-10pt}
\caption{Stochastic results on ETH and UCY in terms of ADE/FDE. $\downarrow$ denotes lower is better.}
\vspace{-12pt}
\label{tab:eth_cvae_results}
\end{table*}
\vspace{-10pt}
\begin{table}[h]
\small
\begin{center}
\begin{tabular}{lcc}
\toprule
\multicolumn{1}{c}{\multirow{2}{*}{Method}} & NuScenes (K=5) &NuScenes (K=10)\\ \cmidrule(lr){2-3} 
\multicolumn{1}{c}{}& \begin{tabular}[c]{@{}l@{}}ADE / FDE (6s) $\downarrow$ \\
\end{tabular} & \begin{tabular}[c]{@{}l@{}}ADE / FDE (6s)  $\downarrow$ \\\end{tabular} \\

\midrule
SGDNet-ED   &2.1 / 4.65 &1.67 / 3.53\\
SGDNet-ED (MI)  & \textbf{1.85} / \textbf{3.87}  &\textbf{1.32} / \textbf{2.50}\\

\bottomrule
\end{tabular}
\end{center}
\vspace{-10pt}
\caption{Results on NuScenes without (first row) and with (second row) map and interaction (MI).}
\vspace{-8pt}
\label{tab:nuscenes}
\vspace{-10pt}
\end{table}

\vspace{5pt}
\xhdr{What if the decoder or the encoder does not receive predicted goals?}
To evaluate the importance of using predicted goals in the decoder and encoder, we implemented two baselines that remove the connection between SGE and the decoder (first row) or the encoder (second row). Table~\ref{tab:ablation_sge} demonstrates that our best model (last row) outperforms both baselines significantly, highlighting the importance of incorporating predicted goals in both the decoder and encoder for achieving more accurate results.

\vspace{-5pt}
\xhdr{Does SGE need supervision?}
To investigate whether supervised stepwise goals offer 
more accurate information for trajectory prediction, 
we train our network without the goal module loss, 
$\mbox{RMSE}(\mathbf{\widehat{Y}}_{t}^g,\mathbf{Y}_{t})$. 
This change decreases the accuracy of our best model by 
13/68/208 and 15/57/181 on 
JAAD/PIE for 0.5s, 1.0s, and 1.5s MSE, respectively. 
This suggests supervision on SGE leads to better goal representation 
to help predicting trajectory, and thus using loss to optimize SGE is 
important in our model.

\vspace{-4pt}
\xhdr{What if we exclude goal aggregator?}
For each time step, each individual goal has a different
impact on the prediction. Thus we develop goal aggregators that use an attention mechanism to understand the relative importance of different subsets of future goals. 
The results in Table~\ref{tab:ablation_sge} show that excluding this attention mechanism (row 3) reduces significantly  reduces results compared to our full model.

\vspace{-3pt}
\xhdr{Are stepwise goals better than fewer goals?}
To further illustrate that using stepwise goals is superior to fewer or only long-term goals, we replace stepwise goals with different number of goals in the encoder and decoder. We show the results in Table~\ref{tab:ablation_sge} (Row 4-9). For each setting, we always include the last goal, and add more goals incrementally. The results indicate that as goals are added, they offer extra information and can significantly enhance trajectory prediction performance. Performance is optimal when goals are predicted at each time step.

Our full model (SGNet-ED) thus uses GRU as the backbone for SGE, and
the output stepwise goals are fed into both encoder and decoder.
Our final loss term includes goal loss, BoM trajectory loss, and KLD loss.

\vspace{-3pt}
\xhdr{How to handle multi-modal data?}
Handling multi-modal input such as social cues/interactions or ego-motion is a classic problem in trajectory prediction. For demonstration, we conducted an experiment on NuScenes prediction challenge split in the Table~\ref{tab:nuscenes} to
show the ability to integrate map and interaction information:
we followed CoverNet~\cite{Phan-Minh_2020_CVPR} to rasterize the location of each agent in the scene and overlay it on the map. We use a four layer CNN to extract features from the overplayed map and combine it with the initial decoder hidden state. We submitted our result to NuScenes and ranked third place in NuScenes prediction challenge at the
6th AI Driving Olympics~\cite{olympics}.
The results suggest that social cues and map information can be implicitly learned and integrated into our model, considerably improving the final results.
Note our proposed module is a simple and efficient temporal unit, and we only take historical trajectory as an input in the following experiments.

\begin{figure*}[t]
\vspace{5pt}
\centering
\includegraphics[width=1\linewidth]{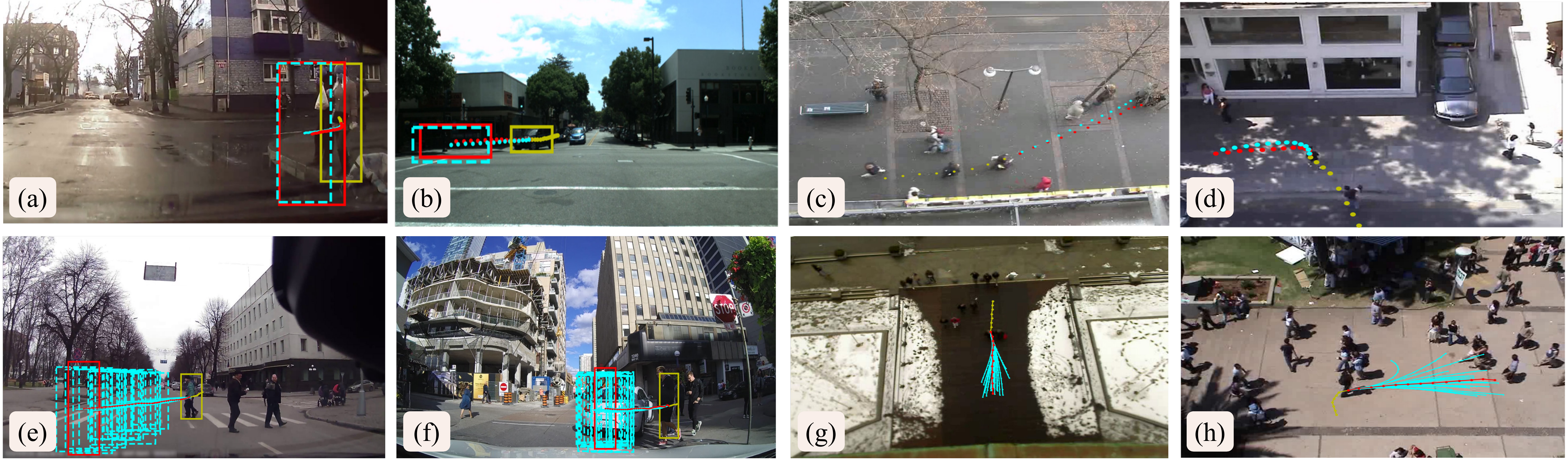}
\vspace{-15pt}
\caption{Qualitative results. The yellow color indicates the observed trajectory, the red color indicates the ground truth future trajectory, and the cyan color indicates the predictions from our SGDNet-ED model (better in color).}
\vspace{-12pt}
\label{fig:qualitative_results}
\end{figure*}

\begin{figure}[t]
\centering
\includegraphics[width=1\linewidth]{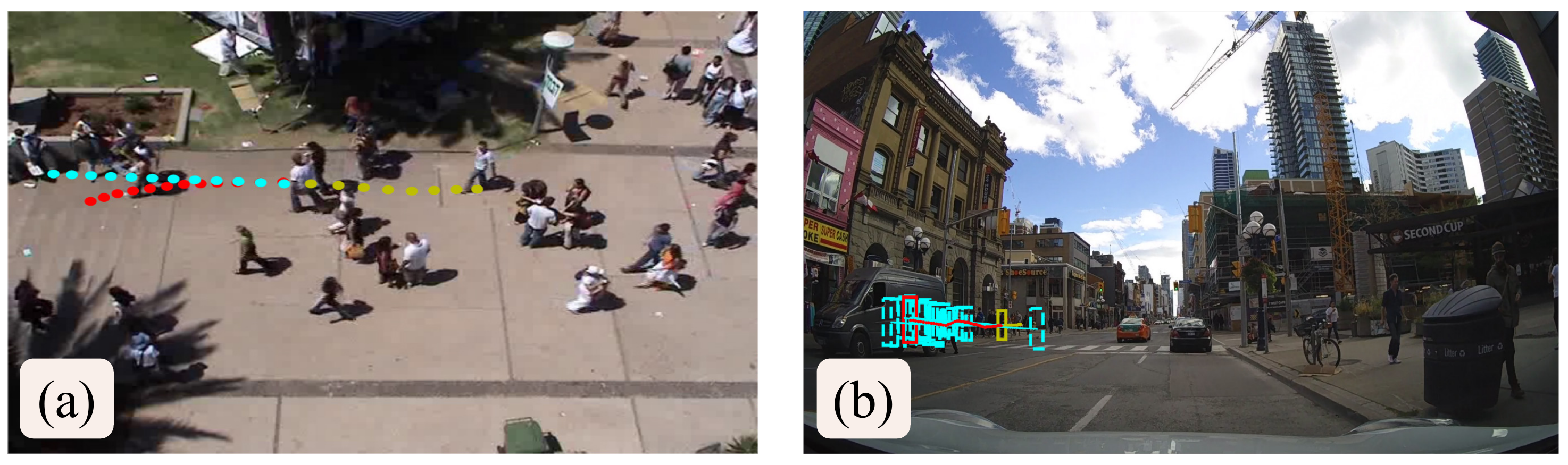}
\vspace{-15pt}
\caption{Failure cases of deterministic (a) and stochastic (b) predictions. The yellow color indicates the observed trajectory, the red color indicates the ground truth future trajectory, and cyan color indicates the predictions from our SGNet-ED model (better in color).}
\vspace{-12pt}
\label{fig:failure_results}
\end{figure}

\vspace{-3pt}
\subsection{Comparison with the State-of-the-art}
\vspace{-3pt}
In this section, we compare our best model under two different
settings: deterministic, in which the model returns a single
trajectory, and stochastic, in which we report the best-performing
sample among $K$ trajectories.

\subsubsection{Deterministic Results on First-person Benchmarks}

We start by evaluating our model's performance on predicting pedestrian trajectory in two first-person datasets.
As shown in Table~\ref{tab:jaadpie_results}, our model (SGNet-ED)
significantly outperforms the state-of-the-art on the first-person
pedestrian detection datasets. 
Compared to~\cite{yao2020bitrap} on JAAD, our model reduces the MSE error by 13\%, 15\%, and 15\% for 0.5s, 1.0s, and 1.5s prediction, respectively. On PIE, our model
reduces MSE  by 3\%, 9\%, and 11\% for 0.5s, 1.0s, and 1.5s
prediction. As the prediction time range increases (and thus the
problem becomes harder), our model decreases MSE  more
significantly, suggesting that SGE is especially helpful for long-term prediction. 
We obtain similar results from other evaluation metrics, as shown in Table~\ref{tab:jaadpie_results}.
In addition, we perform experiments on the HEV-I first-person vehicle dataset. Our SGNet-ED yields 6.28, 11.35 and 18.27 for 0.5s, 1.0s, and 1.5s ADE, 39.86 for FDE, and 0.63 for FIOU. 
Our results improve by an average of 10\% over \cite{yao2018egocentric}.

\subsubsection{Deterministic Results on Third-person Benchmarks}

Table~\ref{tab:eth_results} shows that our model outperforms the
state-of-the-art by more than 10\% in terms of ADE and FDE on
average.

\subsubsection{Stochastic Results on First-person Benchmarks}
To fairly compare with~\cite{yao2020bitrap},
we generate $K=20$ proposals and report the best-performing sample.
For the first-person
datasets, our method outperforms the state-of-the-art by an average of 14\% on
JAAD and 25\% on PIE (Table~\ref{tab:jaadpie_cvae_results}). 

\subsubsection{Stochastic Results on Third-person Benchmarks}
For ETH-UCY, we follow the leave-one-out evaluation protocol with $K = 20$ by following the prior work in
Table~\ref{tab:eth_cvae_results}.  SGNet-ED outperforms the current
state-of-the-art stochastic model~\cite{yao2020bitrap} by 5\% on
average on ETH, ZARA1 and ZARA2, and achieves comparable results on
HOTEL and UNIV.

As with the first-person datasets,
our model leads to a larger improvement as the
prediction length increases, implying that estimated stepwise goals provide better temporal information for accurately predicting the location and magnitude of objects.
For the third-person dataset, our model does not explicitly model interaction, which may explain why our
model is better on less complex scenes (ETH, ZARA1 and ZARA2).
Including interaction or scene maps may help our model to improve
on crowd scenes.

\subsection{Qualitative Results} \label{sec:qualitative}
The first row of Fig.~\ref{fig:qualitative_results} shows four
examples of our best model's deterministic predictions on JAAD, HEV-I, ETH, and UCY, respectively.
In (a), the pedestrian intends to cross the street as the ego-vehicle approaches. The historical trajectory was determined by the pedestrian's movement and the vehicle's ego-motion. The prediction shows the intention of the pedestrian to cross ahead of the ego-vehicle, avoiding a collision. In (b), the target vehicle makes a right turn ahead of the ego-vehicle, and the ego-vehicle waits for it after the turn. In (c), the pedestrian follows a curved route, and (d) demonstrates our ability to notice the pedestrian's immediate change and make an accurate prediction. 

The second row of Fig.~\ref{fig:qualitative_results} shows four examples of our best model's stochastic predictions. 
We show all 20 stochastic trajectories
generated by our best model. Images (e) and (f) illustrate the results of JAAD and PIE, respectively, which both anticipate the presence of a pedestrian crossing the street. Images (g) and (h) show the scene in ETH and UCY.
Most of the predictions are close to the ground truth trajectory and bounding box, indicating the stability of our stochastic prediction model.

\vspace{-5pt}
\subsection{Failure Cases} \label{limitation}
\vspace{-1pt}

Fig.~\ref{fig:failure_results} studies two
failure cases of our model. 
In case (a), the marked person is descending the stairs in the near future, but our model fails to predict the correct trajectory due to the lack of context information and interaction between pedestrians.
In case (b), most of the stochastic predictions show the
pedestrian walking along the road, but one prediction indicates the
possibility that the pedestrian may cross the road along the zebra
crossing. However, this error may be a ``blessing in disguise'', for example helping an autonomous driving system to prepare for a potential future risk.

\vspace{-2pt}
\section{Conclusion}
We presented SGNet to
tackle the trajectory prediction problem.
Unlike most existing goal-driven models that only
estimates final destination or distant goals, SGNet
predicts both long- and short-term goals and explicitly incorporates
these estimated future states.
We showed that these goals can help to predict the trajectory. 
We conducted extensive experiments on three first-person and three bird's eye view
datasets to evaluate the
proposed approach. Experimental results showed the effectiveness and robustness of our models against the state-of-the-art methods.


\bibliographystyle{IEEEtran}
\bibliography{egbib}

\end{document}